\documentclass[sigconf]{acmart}
\usepackage{multirow}

\AtBeginDocument{%
  }

\setcopyright{acmlicensed}
\copyrightyear{2024}
\acmYear{2024}
\acmDOI{XXXXXXX.XXXXXXX}

\acmConference[KDDCup'24]{the 30th ACM SIGKDD International Conference on Knowledge Discovery and Data Mining.}{Aug 25--29, 2024}{Barcelona, Spain}
\acmISBN{978-1-4503-XXXX-X/18/06}

\begin{document}

\title{EC-Guide: A Comprehensive E-Commerce Guide for Instruction Tuning and Quantization}

\author{Zhaopeng Feng}
\authornote{Equal contribution.}
\author{Zijie Meng}
\authornotemark[1]
\author{Zuozhu Liu}
\authornote{Corresponding author.}
\affiliation{%
  \institution{Zhejiang University-University of Illinois at Urbana Champaign Institute, Zhejiang University}
  \city{Jiaxing}
  \state{Zhejiang}
  \country{China}
}
\email{{zhaopeng.23, zijie.22, zuozhuliu}@intl.zju.edu.cn}

\renewcommand{\shortauthors}{Zhaopeng Feng, Zijie Meng, Zuozhu Liu}

\begin{abstract}
    Large language models (LLMs) have attracted considerable attention in various fields for their cost-effective solutions to diverse challenges, especially with advancements in instruction tuning and quantization. E-commerce, with its complex tasks and extensive product-user interactions, presents a promising application area for LLMs. However, the domain-specific concepts and knowledge inherent in e-commerce pose significant challenges for adapting general LLMs. To address this issue, we developed EC-Guide\footnote{\href{https://github.com/fzp0424/EC-Guide-KDDUP-2024}{https://github.com/fzp0424/EC-Guide-KDDUP-2024}}, a comprehensive e-commerce guide for instruction tuning and quantization of LLMs. We also heuristically integrated Chain-of-Thought (CoT) during inference to enhance arithmetic performance. Our approach achieved the 2nd place in Track 2 and 5th place in Track 5 at the Amazon KDD Cup'24\footnote{\href{https://www.aicrowd.com/challenges/amazon-kdd-cup-2024-multi-task-online-shopping-challenge-for-llms}{https://www.aicrowd.com/challenges/amazon-kdd-cup-2024}}. Additionally, our solution is model-agnostic, enabling effective scalability across larger systems.
\end{abstract}




\maketitle

\section{Introduction}
Current techniques struggle to understand the nuances of specific shopping terms, customer behaviors, preferences, and the diverse array of products and languages~\citep{ecinstruct}. With the advent of large language models (LLMs), there is a growing belief in their capability to tackle these challenges. To this end, the organizers of Amazon KDD Cup'24 introduced ShopBench, a benchmark designed to simulate the complexities of online shopping. It includes 57 tasks and approximately 20,000 questions sourced from real-world Amazon shopping data. The competition includes 5 Tracks: 
\begin{itemize}
    \item Shopping Concept Understanding: Decoding complex shopping concepts and terminologies.
    \item Shopping Knowledge Reasoning: Making informed decisions based on shopping knowledge. 
    \item User Behavior Alignment: Understanding dynamic customer behavior. 
    \item Multilingual Abilities: Shopping across languages.
    \item All-Around: Solving all questions with a single solution.
\end{itemize}
Our team ``ZJU-AI4H'' finally achieved 2nd place in Track 2 and 5th place in Track 5. Our solution for both Tracks can be summarized as three key steps: dataset construction, instruction tuning, and post training quantization. Especially, we noticed that Chain-of-Thought (CoT)~\citep{cot} can significantly boost the arithmetic performance.

\begin{figure*}[ht]
  \centering
  \includegraphics[width=\linewidth]{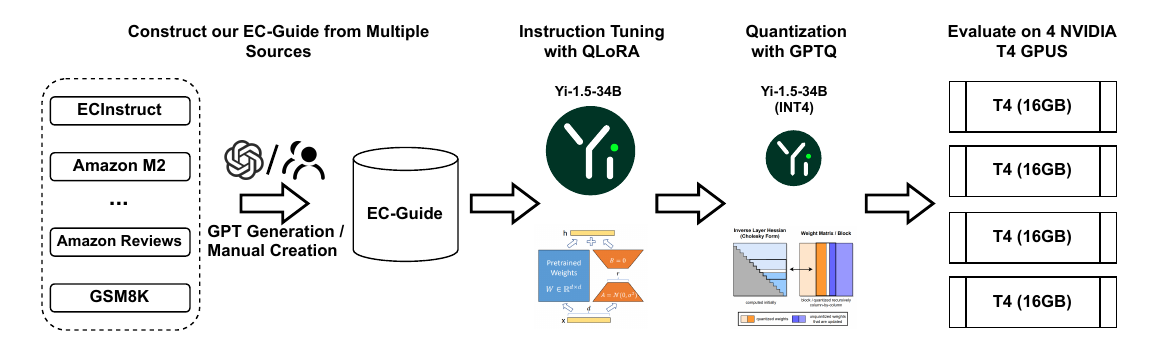}
  \caption{Illustration of our solution. We first construct our EC-Guide (74k examples for 24 sub-tasks across 5 types) dataset from multiple sources~\citep{ecinstruct,amazon-m2}. Then we finetune Yi-1.5-34B~\citep{young2024yi} with QLoRA~\citep{qlora} and quantize it with GPTQ~\citep{frantar2022gptq}.
}
 \label{pipeline}
\end{figure*}

\section{Related Works}
As a data-driven technology, LLMs exhibit exceptional performance, promoting the development of numerous datasets. GSM8K~\citep{gsm8k} focuses on grade school mathematics requiring multi-step reasoning. ECInstruct~\citep{ecinstruct} introduces diverse e-commerce subtasks to guide the instruction tuning of general LLMs. Amazon-M2~\citep{amazon-m2} is a multilingual product recommendation dataset enhancing the understanding of user preferences. Additionally, other datasets~\citep{commonsenseqa, esci, amazonqa, PairwiseFashion, IntentionQA, ner_Indonesian, Womens_Clothing_Reviews} provide varied formats for evaluating LLMs. To further explore e-commerce applications, we developed EC-Guide, a comprehensive guide for instruction tuning and quantization.

Apart from the rapid development of datasets, researchers are increasingly focusing on training and inference techniques. LoRA~\citep{lora} significantly reduces fine-tuning costs by freezing LLM's weights and injecting a learnable matrix. QLoRA~\citep{qlora} further reduces costs by introducing quantization techniques. GPTQ~\citep{frantar2022gptq} is a one-shot weight quantization for LLMs by leveraging approximate second-order information to compress models down to lower bits
(like 4-bit) per weight with minimal impact on performance. CoT~\citep{cot} enhances the reasoning ability of LLMs through appropriate prompts without additional training. Based on these methods, we adapted general LLM to specific domains through instruction tuning and deployed by quantization and CoT within limited resources.

\section{Methodology}
We first constructed our instruction tuning dataset, EC-Guide, by mining existing datasets~\citep{ecinstruct, amazon-m2}, and utilizing ChatGPT\footnote{\href{https://openai.com/chatgpt/}{https://openai.com/chatgpt/}} for data generation. We also expanded the official development dataset to enhance evaluation. Through instruction tuning, we embedded knowledge specifically relevant to e-commerce tasks into existing LLMs. However, in Round 2, solutions had access to 4$\times$NVIDIA T4 (16GB) GPUs\footnote{\href{https://www.nvidia.cn/data-center/tesla-t4/}{https://www.nvidia.cn/data-center/tesla-t4/}}, making it impractical to directly deploy powerful LLMs like Yi-1.5-34B~\citep{young2024yi} or Qwen2-72B~\citep{yang2024qwen2} without quantization. During inference, we applied CoT to further boost LLM calculation performance. Details are as follows.

\subsection{Dataset Construction}
Due to the lack of large-scale training datasets for the Amazon KDD Cup'24, we employed paraphrasing and extending existing datasets~\citep{ecinstruct, amazon-m2} to develop EC-Guide, encompassing 74k examples across five task types: Generation, Multiple Choice Question, Retrieval, Ranking, and Named Entity Recognition. Statistic details are shown in Table~\ref{tab:dataset}. Furthermore, we expanded the official development dataset from 96 to 506 examples by strategies such as option reordering and segment sampling from our EC-Guide, which significantly enhances the robustness of the development dataset to effectively evaluate our model.

\begin{table}[t]
\centering
\caption{Statistics of EC-Guide.}
\label{tab:dataset}
\resizebox{\columnwidth}{!}{%
\begin{tabular}{llll} 
\toprule
\textbf{\textbf{\textbf{\textbf{Task type}}}} & \textbf{Sub-tasks} & \textbf{\# of examples} & \textbf{\textbf{\textbf{\textbf{Source}}}}  \\ 
\midrule
\multirow{10}{*}{Generation} & Product Elaboration (PE) & 479 & \cite{ecinstruct} \\
{} & Product Question and Answer (PQA) & 6,834 & \cite{amazonqa} \\
{} & Category Recognition (CR) & 1,000 & \cite{amazonmetadata} \\
{} & Explaining Pair Fashion (EPF) & 3,000 & \cite{PairwiseFashion} \\
{} & Explaining Bought Together (EBT) & 2,315 & \cite{IntentionQA} \\
{} & Extract Review Keyphrase (ERK) & 1,000 & \cite{ecinstruct} \\
{} & Extract Product Keyphrase (EPK) & 3,000 & \cite{PairwiseFashion} \\
{} & Product Keyword Summarization (PKS) & 1,296 & \cite{esci, ecinstruct, amazonreview} \\ 
{} & Review Title Summarization (RTS) & 1,455 & \cite{amazonreview, Womens_Clothing_Reviews} \\
{} & Multilingual Translation (MT) & 2,997 & \cite{amazon-m2, flores} \\
\midrule
\multirow{9}{*}{Multiple Choice Question } & Select Product based on Attribute (SPA) & 520 & \cite{ecinstruct} \\
 & Select Attribute based on Product (SAP) & 1,385 & \cite{amazonreview} \\
 & Product Relation Prediction (PRP) & 1,499 & \cite{ecinstruct} \\
 & Query Product Relation Prediction (QPRP) & 2,150 & \cite{esci} \\
 & Query Product Relation Judgement (QPRJ) & 501 & \cite{ecinstruct} \\
 & Sentiment Analysis (SA) & 3,500 & \cite{ecinstruct, Womens_Clothing_Reviews} \\
 & Product Keyword Summarization (PKS) & 271 & \cite{esci} \\
 & Multilingual Description Matching (MDM) & 500 & \cite{amazonreview} \\
 & Arithmetic and Commonsense Reasoning (ACR) & 7,184 & \cite{gsm8k, commonsenseqa} \\
\midrule
\multirow{5}{*}{Retrieval} & Inferring Potential Purchases (IPP) & 10,774 & \cite{ecinstruct, amazon-m2} \\
 & Retrieving Review Snippets (RRS) & 810 & \cite{amazonreview} \\
 & Retrieving Review Aspects (RRA) & 1,000 & \cite{amazonreview} \\ 
 & Category Recognition (CR) & 7,500 & \cite{amazonmetadata} \\
 & Product Recognition (PR) & 2,297 & \cite{amazonmetadata} \\
\midrule
Ranking & Query Product Ranking (QPR) & 4,008 & \cite{esci} \\
\midrule
Named Entity Recognition & Named Entity Recognition (NER) & 7,429 & \cite{ecinstruct, amazonreview, ner_Indonesian} \\
\midrule
ALL & - & 74,704 & - \\
\bottomrule
\end{tabular}
}
\end{table}
\subsubsection{Generation} \label{sec:generation}
In the generation task, each question is paired with a specific instruction, and the model is to generate text that precisely follows these instructions. There are multiple types of generation questions, including elaboration, extraction, summarization and translation.

For the elaboration set, we introduced the following subtasks:
1. Product Elaboration (PE): We utilized the ``Attribute Value Extraction'' from ECInstruct~\citep{ecinstruct} to collect type-related attributes, then filtered to obtain 479 entities, and used ChatGPT to generate detailed descriptions for each entity. 
2. Product Question and Answer (PQA): We selected samples from \citet{amazonqa} with only one answer, filtered them based on answer length, and finally obtained 6,834 QA pairs.
3. Category Recognition (CR): We extracted product-category relationships from \citet{amazonmetadata}, and sampled 1,000 entries to prompt the model to recognize categories from product titles.
4. Explaining Pairwise Fashion (EPF): We modified prompts from the PFE dataset~\citep{PairwiseFashion} and filtered based on response length, yielding 3,000 instances to evaluate whether the mentioned clothes matched.
5. Explaining Bought Together (EBT): We directly integrated 2,315 entries from IntentionQA~\citep{IntentionQA} to assess the reasons for buying two products together.

For the extraction set, we introduced two subtasks: extracting review keyphrases (ERK) and extracting product keyphrases (EPK), which focus on extracting keyphrases related to aspects or features from reviews and products, respectively.
ERK is based on the ``Sentiment Analysis'' of ECInstruct~\citep{ecinstruct}, where we used ChatGPT to extract aspect-keyphrase pairs from reviews and obtained 1,000 samples. 
EPK is sourced from the PFE dataset~\citep{PairwiseFashion}, where we filtered out data items with overly short descriptions and extracted feature-keyphrase pairs, yielding 3,000 cases.

The summarization set includes two subtasks:
1. Product Keyword Summarization (PKS) involves summarizing keywords list to encapsulate product information. We sampled product information from existing datasets~\citep{esci, ecinstruct, amazonreview}, and used ChatGPT for annotation to produce 1,296 entries.
2. Review Title Summarization (RTS) aims to create concise titles for reviews. We filtered previous dataset~\citep{amazonreview, Womens_Clothing_Reviews} to obtain 1,455 cases.

For our translation set, we employed ChatGPT to translate product titles between English and several other languages, including Spanish, German, Italian, Japanese, and French. We also translated from other languages into English, resulting in a total of 2,000 translation pairs. Additionally, we utilized samples from Flores~\citep{flores} to enhance our translation tasks, resulting in 997 additional pairs.

\subsubsection{Multiple Choice Question}  \label{sec:mcq}
Multiple choice question, widely used for their objective assessment, require selecting the correct answer from the choices list identified by Arabic numerals. Specifically, we include the following subtasks:
1. Select Product based on Attribute (SPA): We sampled attributes from the ``Attribute Value Extraction'' of ECInstruct~\citep{ecinstruct}, and generated product options, resulting in 520 data entries.
2. Select Attribute based on Product (SAP): Using product titles extracted from Amazon Reviews~\citep{amazonreview}, we generated multiple choice questions about attributes with ChatGPT, yielding 1,385 data entries. 
3. Product Relation Prediction (PRP): We extracted 1,499 cases from ``Product Relation Prediction'' and ``Product Matching'' in ECInstruct~\citep{ecinstruct}. The former predicts potential purchase or browsing intentions between two products, while the latter judges whether two products are the same.
4. Query Product Relation Prediction (QPRP): Based on relationships between queries or products in the Shopping Queries Dataset~\citep{esci}, we constructed 2,150 cases.
5. Query Product Relation Judgement (QPRJ): Sampled from ``Product Substitute Indentification'' of ECInstruct~\citep{ecinstruct}, QPRJ includes 501 examples.
6. Sentiment Analysis (SA): We directly extracted 3,500 data entries from ``Sentiment Analysis'' of ECInstruct~\citep{ecinstruct} and Women's Clothing E-Commerce Reviews~\citep{Womens_Clothing_Reviews}.
7. Product Keyword Summarization (PKS): Using a method similar to PKS in Section~\ref{sec:generation}, we obtained 271 cases.
8. Multilingual Description Matching (MDM): MDM comprises 300 cases, aiming to match product titles with the correct features in multiple languages. We filtered products from Amazon Reviews~\citep{amazonreview} and translated their features into various languages (English, Spanish, German, Italian, Japanese, and French) with ChatGPT. And then we sampled three features from other products to form the options for each final question.
9. Arithmetic and Commonsense Reasoning (ACR): We obtained 7,184 items from GSM8K~\citep{gsm8k} and CommonsenseQA~\citep{commonsenseqa}. And we annotated the rationale with ChatGPT for CommonsenseQA.

\subsubsection{Retrieval}  \label{sec:retrieval}
In this task, the model's objective is to retrieve answers from a list of candidate items to meet specific requirements. The subtasks are as follows:
1. Inferring Potential Purchases (IPP): There are two main categories multi-to-one and one-to-multi in IPP. For the multi-to-one, which predicts the next purchase item based on multiple items in the purchase history, we filtered cases from the ``Sequential Recommendation'' in ECInstruct~\citep{ecinstruct} to obtain 3,950 cases. For the one-to-multi, which predicts multiple potential purchase items based on a single purchased item, we mined the data from Amazon-M2~\citep{amazon-m2}, resulting in a total of 6,824 cases.
2. Retrieving Review Snippets (RRS): We sampled Amazon Reviews~\citep{amazonreview} to obtain 3,000 products and their corresponding reviews. Then we generated multiple aspect-snippet pairs from the reviews with ChatGPT. Finally, considering the proportion of positive and negative reviews, we created a total of 810 data entries.
3. Retrieving Review Aspects (RRA): Using the same data source as RRS, we randomly combined reviews and aspects to obtain 1,000 data entries.
4. Category Recognition (CR): Similar to the CR in Section~\ref{sec:generation}, we constructed retrieval lists by randomly sampling categories, resulting in 7,500 data items.
5. Product Recognition (PR): Using the data from CR, we reversed the product-category table to create a category-product dictionary, resulting in 2,297 data entries.

\subsubsection{Ranking} \label{ranking}
In this task, the model's goal is to reorganize items in the candidate list based on how well they meet the requirements. We utilized the Shopping Queries Dataset~\citep{esci}, which assesses the relevance between the queries and products using ESCI judgments (Exact, Substitute, Complement, Irrelevant). Furthermore, leveraging the multilingual source dataset, our ranking instances comprises 2,064 queries in English, 790 in Japanese, and 1,184 in Spanish. 

\subsubsection{Named Entity Recognition} \label{sec:ner}
Named Entity Recognition (NER) is an important benchmark for evaluating LLMs and is prevalent across various domains. In this task, we extracted 1,446 entries from ``Attribute Value Extraction'' of ECInstruct~\citep{ecinstruct}, 1,099 entries from Amazon Reviews~\citep{amazonreview}, and 4,884 entries from \citet{ner_Indonesian}.

\subsection{Instruction Tuning}
Instruction tuning enhances the model's ability to generate contextually appropriate answers based on the given instruction, which allows the model to understand and execute tasks specified in the instructions. During training, the objective is to optimize the model ${\mathcal{\pi}_\theta}$ using a curated dataset $\mathcal{D}_{IT}=\{x^{(i)}, y^{(i)}\}_{i=1}^{N}$ by minimizing the negative log-likelihood of the output $y$:
\begin{equation}
     \mathcal{L}(\theta; \mathcal{D}_{IT}) = -\mathbb{E}_{(x, y) \sim \mathcal{D}_{IT}} \left[\log \pi_{\theta}(y \mid x)\right]
\end{equation}
Before tuning, we organized raw data in [\textit{Instruction}, \textit{Input}, \textit{Output}], concatenated them, and applied chat template to form $\mathcal{D}_{IT}$. 

Deploying models with larger parameters necessitates effective compression methods, such as model quantization, to accelerate inference and save memory. Quantization involves converting high-precision parameters to lower-precision formats without altering the model's parameters or architecture.

\subsection{Post Training Quantization}
It is widely accepted that larger scale generally leads to greater model capabilities~\citep{yang2024qwen2, young2024yi}. However, LLMs typically store parameters in high-precision floating-point formats, demanding significant computational resources for inference. Deploying larger models necessitates effective compression methods, such as quantization, to accelerate inference and save memory. Therefore, we utilize GPTQ~\citep{frantar2022gptq}, which is a training-free quantization for LLMs to achieve high accuracy and efficiency. We detailed the memory of weights loaded in Table~\ref{tab:multi}. In particular, we equally sample training examples based on task types from our training set, ultimately allocate 1,846 examples for quantization. Additionally, we also compare our quantization with samples from the external data source like C4\footnote{\href{https://huggingface.co/datasets/allenai/c4}{https://huggingface.co/datasets/allenai/c4}} as shown in Table~\ref{tab:multi}.

\subsection{Chain-of-Thought (CoT) Reasoning}
After instruction tuning and quantization on the elaborately designed EC-Guide, we introduced CoT~\citep{cot} in the inference only for multiple choice questions involving calculations. Specifically, we employed a heuristic strategy to determine if a question belong to arithmetic-based by counting the number of digits. We then prefixed the prompt with ``Let's think step by step.'', stimulating the model to generate a rationale that leads to the correct answer. Notably, we observed that applying CoT to Track 2 in Round 1 increased the score from 0.7417 to 0.7908 with the same model.


\begin{table}[t]
    \centering
    \caption{Results of various models under the same training settings. The loading weights are averaged across 4 GPU.}
    \label{tab:multi}
    \resizebox{\linewidth}{!}{%
    \begin{tabular}{ccccccc}
        \toprule
        \multirow{2.5}{*}{\textbf{Model}} & \multirow{2.5}{*}{\textbf{\begin{tabular}{c}PTQ\\(dataset)\end{tabular}}} & \multirow{2.5}{*}{\textbf{\begin{tabular}{c}Loading\\Weights\end{tabular}}} & \multicolumn{2}{c}{\textbf{Track 2}} & \multicolumn{2}{c}{\textbf{Track 5}} \\ 
        \cmidrule(r{1mm}){4-5} \cmidrule(l{1mm}){6-7}
        {} & {} & {} & \textbf{Dev} & \textbf{Round 2} & \textbf{Dev} & \textbf{Round 2} \\ 
        \midrule
        {Qwen2-7B} & {\textbackslash{}} & {3.55 GB} & {0.6562} & {0.6706} & {0.7092} & {0.7224} \\
        {Yi-1.5-9B} & {\textbackslash{}} & {4.45 GB} & {0.7031} & {0.6628} & {0.7153} & {0.7048} \\
        {GLM-4-9B} & {\textbackslash{}} & {6.65 GB} & {0.7187} & {0.6875} & {0.7299} & {0.7163} \\
        \multirow{3}{*}{Yi-1.5-34B} & {INT4 (C4)} & {4.55 GB} & {0.6562} & {-} & {0.7017} & {-} \\
        {} & {INT4 (Ours)} & {4.55 GB} & {0.7500} & {0.7452} & {0.7403} & {0.7323} \\
        {} & {\textbackslash{}} & {16.04 GB} & {0.7968} & {-} & {0.7583} & {-} \\
        \bottomrule
    \end{tabular}%
    }
\end{table}

\begin{table}[t]
\centering
\caption{Results of submission with Yi-1.5-34B.}
\label{tab:yi_result}
\resizebox{\columnwidth}{!}{%
\begin{tabular}{cccc} 
\toprule
\textbf{\textbf{\textbf{\textbf{Model}}}}                                   & \textbf{Training Set (ratio, \# of cases)}               & \multicolumn{1}{l}{\textbf{\textbf{Track 2}}} & \textbf{Track 5}  \\ 
\midrule
\multirow{3}{*}{\begin{tabular}[c]{@{}c@{}}Yi-1.5-34B\\(INT4)\end{tabular}} & $\lambda$1: $\lambda$2: $\lambda$3: $\lambda$4: $\lambda$5 = 0.31: 0.23: 0.30: 0.05: 0.10, 74704 & 0.7487                                        & \textbf{0.7462}   \\
                                                                            & $\lambda$1: $\lambda$2: $\lambda$3: $\lambda$4: $\lambda$5 = 0.33: 0.40: 0.17: 0.02: 0.08, 15764 & \textbf{0.7834}                               & 0.7386            \\
                                                                            & $\lambda$1: $\lambda$2: $\lambda$3: $\lambda$4: $\lambda$5 = 0.46: 0.24: 0.14: 0.04: 0.12, 15043 & 0.7440                                        & 0.7402            \\
\bottomrule
\end{tabular}
}
\end{table}

\section{Experiments}
We finetuned all models on 4$\times$A40 GPUs or 8$\times$RTX3090 GPUs with QLoRA. We deployed LLMs by vllm\footnote{\href{https://github.com/vllm-project/vllm}{https://github.com/vllm-project/vllm}}, which utilizes PagedAttention to manage attention keys and values, to accelerate inference.

Table~\ref{tab:multi} demonstrates the performance of different models with the same training setting. Notably, Yi-1.5-34B achieved the highest scores across both Track 2 and 5 in both development and official test set. We also observed that models quantized using out-of-domain datasets C4 exhibited significant performance drops compared to those using in-domain sampled data. Table~\ref{tab:yi_result} presents our ablation study, which highlights the influence of different ratios of task types in training set, and suggests that smaller training sets sometimes outperform larger ones in specific scenarios. This observation leads us to hypothesize about the existence of a performance trade-off among different tasks.

\section{Conclusion}
The Amazon KDD Cup’24 competition presents a unique challenge by focusing on the application of LLMs in E-commerce across multiple tasks. Our solution for addressing Tracks 2 and 5 involves a comprehensive pipeline encompassing dataset construction, instruction tuning and post-training quantization. The core of our strategy is EC-Guide specifically tailored for E-commerce scenarios. Notably, we heuristically integrated CoT reasoning to enhance the arithmetic capabilities of LLMs, resulting in improved performance in both Tracks.


\bibliographystyle{ACM-Reference-Format}
\bibliography{reference}


\begin{thebibliography}{19}


\ifx \showCODEN    \undefined \def \showCODEN     #1{\unskip}     \fi
\ifx \showDOI      \undefined \def \showDOI       #1{#1}\fi
\ifx \showISBNx    \undefined \def \showISBNx     #1{\unskip}     \fi
\ifx \showISBNxiii \undefined \def \showISBNxiii  #1{\unskip}     \fi
\ifx \showISSN     \undefined \def \showISSN      #1{\unskip}     \fi
\ifx \showLCCN     \undefined \def \showLCCN      #1{\unskip}     \fi
\ifx \shownote     \undefined \def \shownote      #1{#1}          \fi
\ifx \showarticletitle \undefined \def \showarticletitle #1{#1}   \fi
\ifx \showURL      \undefined \def \showURL       {\relax}        \fi
\providecommand\bibfield[2]{#2}
\providecommand\bibinfo[2]{#2}
\providecommand\natexlab[1]{#1}
\providecommand\showeprint[2][]{arXiv:#2}

\bibitem[Cobbe et~al\mbox{.}(2021)]%
        {gsm8k}
\bibfield{author}{\bibinfo{person}{Karl Cobbe}, \bibinfo{person}{Vineet Kosaraju}, \bibinfo{person}{Mohammad Bavarian}, \bibinfo{person}{Mark Chen}, \bibinfo{person}{Heewoo Jun}, \bibinfo{person}{Lukasz Kaiser}, \bibinfo{person}{Matthias Plappert}, \bibinfo{person}{Jerry Tworek}, \bibinfo{person}{Jacob Hilton}, \bibinfo{person}{Reiichiro Nakano}, \bibinfo{person}{Christopher Hesse}, {and} \bibinfo{person}{John Schulman}.} \bibinfo{year}{2021}\natexlab{}.
\newblock \showarticletitle{Training Verifiers to Solve Math Word Problems}.
\newblock \bibinfo{journal}{\emph{arXiv preprint arXiv:2110.14168}} (\bibinfo{year}{2021}).
\newblock


\bibitem[Dettmers et~al\mbox{.}(2024)]%
        {qlora}
\bibfield{author}{\bibinfo{person}{Tim Dettmers}, \bibinfo{person}{Artidoro Pagnoni}, \bibinfo{person}{Ari Holtzman}, {and} \bibinfo{person}{Luke Zettlemoyer}.} \bibinfo{year}{2024}\natexlab{}.
\newblock \showarticletitle{Qlora: Efficient finetuning of quantized llms}.
\newblock \bibinfo{journal}{\emph{Advances in Neural Information Processing Systems}}  \bibinfo{volume}{36} (\bibinfo{year}{2024}).
\newblock


\bibitem[Ding et~al\mbox{.}(2024)]%
        {IntentionQA}
\bibfield{author}{\bibinfo{person}{Wenxuan Ding}, \bibinfo{person}{Weiqi Wang}, \bibinfo{person}{Sze Heng~Douglas Kwok}, \bibinfo{person}{Minghao Liu}, \bibinfo{person}{Tianqing Fang}, \bibinfo{person}{Jiaxin Bai}, \bibinfo{person}{Junxian He}, {and} \bibinfo{person}{Yangqiu Song}.} \bibinfo{year}{2024}\natexlab{}.
\newblock \showarticletitle{IntentionQA: A Benchmark for Evaluating Purchase Intention Comprehension Abilities of Language Models in E-commerce}.
\newblock \bibinfo{journal}{\emph{arXiv preprint arXiv:2406.10173}} (\bibinfo{year}{2024}).
\newblock


\bibitem[Frantar et~al\mbox{.}(2022)]%
        {frantar2022gptq}
\bibfield{author}{\bibinfo{person}{Elias Frantar}, \bibinfo{person}{Saleh Ashkboos}, \bibinfo{person}{Torsten Hoefler}, {and} \bibinfo{person}{Dan Alistarh}.} \bibinfo{year}{2022}\natexlab{}.
\newblock \showarticletitle{Gptq: Accurate post-training quantization for generative pre-trained transformers}.
\newblock \bibinfo{journal}{\emph{arXiv preprint arXiv:2210.17323}} (\bibinfo{year}{2022}).
\newblock


\bibitem[Goyal et~al\mbox{.}(2022)]%
        {flores}
\bibfield{author}{\bibinfo{person}{Naman Goyal}, \bibinfo{person}{Cynthia Gao}, \bibinfo{person}{Vishrav Chaudhary}, \bibinfo{person}{Peng-Jen Chen}, \bibinfo{person}{Guillaume Wenzek}, \bibinfo{person}{Da Ju}, \bibinfo{person}{Sanjana Krishnan}, \bibinfo{person}{Marc’Aurelio Ranzato}, \bibinfo{person}{Francisco Guzm{\'a}n}, {and} \bibinfo{person}{Angela Fan}.} \bibinfo{year}{2022}\natexlab{}.
\newblock \showarticletitle{The flores-101 evaluation benchmark for low-resource and multilingual machine translation}.
\newblock \bibinfo{journal}{\emph{Transactions of the Association for Computational Linguistics}}  \bibinfo{volume}{10} (\bibinfo{year}{2022}), \bibinfo{pages}{522--538}.
\newblock


\bibitem[Hou et~al\mbox{.}(2024)]%
        {amazonreview}
\bibfield{author}{\bibinfo{person}{Yupeng Hou}, \bibinfo{person}{Jiacheng Li}, \bibinfo{person}{Zhankui He}, \bibinfo{person}{An Yan}, \bibinfo{person}{Xiusi Chen}, {and} \bibinfo{person}{Julian McAuley}.} \bibinfo{year}{2024}\natexlab{}.
\newblock \showarticletitle{Bridging language and items for retrieval and recommendation}.
\newblock \bibinfo{journal}{\emph{arXiv preprint arXiv:2403.03952}} (\bibinfo{year}{2024}).
\newblock


\bibitem[Hu et~al\mbox{.}(2021)]%
        {lora}
\bibfield{author}{\bibinfo{person}{Edward~J Hu}, \bibinfo{person}{Yelong Shen}, \bibinfo{person}{Phillip Wallis}, \bibinfo{person}{Zeyuan Allen-Zhu}, \bibinfo{person}{Yuanzhi Li}, \bibinfo{person}{Shean Wang}, \bibinfo{person}{Lu Wang}, {and} \bibinfo{person}{Weizhu Chen}.} \bibinfo{year}{2021}\natexlab{}.
\newblock \showarticletitle{Lora: Low-rank adaptation of large language models}.
\newblock \bibinfo{journal}{\emph{arXiv preprint arXiv:2106.09685}} (\bibinfo{year}{2021}).
\newblock


\bibitem[Jin et~al\mbox{.}(2024)]%
        {amazon-m2}
\bibfield{author}{\bibinfo{person}{Wei Jin}, \bibinfo{person}{Haitao Mao}, \bibinfo{person}{Zheng Li}, \bibinfo{person}{Haoming Jiang}, \bibinfo{person}{Chen Luo}, \bibinfo{person}{Hongzhi Wen}, \bibinfo{person}{Haoyu Han}, \bibinfo{person}{Hanqing Lu}, \bibinfo{person}{Zhengyang Wang}, \bibinfo{person}{Ruirui Li}, {et~al\mbox{.}}} \bibinfo{year}{2024}\natexlab{}.
\newblock \showarticletitle{Amazon-m2: A multilingual multi-locale shopping session dataset for recommendation and text generation}.
\newblock \bibinfo{journal}{\emph{Advances in Neural Information Processing Systems}}  \bibinfo{volume}{36} (\bibinfo{year}{2024}).
\newblock


\bibitem[Kojima et~al\mbox{.}(2022)]%
        {cot}
\bibfield{author}{\bibinfo{person}{Takeshi Kojima}, \bibinfo{person}{Shixiang~Shane Gu}, \bibinfo{person}{Machel Reid}, \bibinfo{person}{Yutaka Matsuo}, {and} \bibinfo{person}{Yusuke Iwasawa}.} \bibinfo{year}{2022}\natexlab{}.
\newblock \showarticletitle{Large language models are zero-shot reasoners}.
\newblock \bibinfo{journal}{\emph{Advances in neural information processing systems}}  \bibinfo{volume}{35} (\bibinfo{year}{2022}), \bibinfo{pages}{22199--22213}.
\newblock


\bibitem[McAuley et~al\mbox{.}(2015)]%
        {amazonmetadata}
\bibfield{author}{\bibinfo{person}{Julian McAuley}, \bibinfo{person}{Christopher Targett}, \bibinfo{person}{Qinfeng Shi}, {and} \bibinfo{person}{Anton Van Den~Hengel}.} \bibinfo{year}{2015}\natexlab{}.
\newblock \showarticletitle{Image-based recommendations on styles and substitutes}. In \bibinfo{booktitle}{\emph{Proceedings of the 38th international ACM SIGIR conference on research and development in information retrieval}}. \bibinfo{pages}{43--52}.
\newblock


\bibitem[McAuley and Yang(2016)]%
        {amazonqa}
\bibfield{author}{\bibinfo{person}{Julian McAuley} {and} \bibinfo{person}{Alex Yang}.} \bibinfo{year}{2016}\natexlab{}.
\newblock \showarticletitle{Addressing complex and subjective product-related queries with customer reviews}. In \bibinfo{booktitle}{\emph{Proceedings of the 25th International Conference on World Wide Web}}. \bibinfo{pages}{625--635}.
\newblock


\bibitem[nicapotato(2018)]%
        {Womens_Clothing_Reviews}
\bibfield{author}{\bibinfo{person}{nicapotato}.} \bibinfo{year}{2018}\natexlab{}.
\newblock \bibinfo{title}{Women's E-Commerce Clothing Reviews}.
\newblock
\newblock
\urldef\tempurl%
\url{https://www.kaggle.com/datasets/nicapotato/womens-ecommerce-clothing-reviews}
\showURL{%
\tempurl}


\bibitem[Peng et~al\mbox{.}(2024)]%
        {ecinstruct}
\bibfield{author}{\bibinfo{person}{Bo Peng}, \bibinfo{person}{Xinyi Ling}, \bibinfo{person}{Ziru Chen}, \bibinfo{person}{Huan Sun}, {and} \bibinfo{person}{Xia Ning}.} \bibinfo{year}{2024}\natexlab{}.
\newblock \showarticletitle{eCeLLM: Generalizing Large Language Models for E-commerce from Large-scale, High-quality Instruction Data}.
\newblock \bibinfo{journal}{\emph{arXiv preprint arXiv:2402.08831}} (\bibinfo{year}{2024}).
\newblock


\bibitem[Reddy et~al\mbox{.}(2022)]%
        {esci}
\bibfield{author}{\bibinfo{person}{Chandan~K Reddy}, \bibinfo{person}{Llu{\'\i}s M{\`a}rquez}, \bibinfo{person}{Fran Valero}, \bibinfo{person}{Nikhil Rao}, \bibinfo{person}{Hugo Zaragoza}, \bibinfo{person}{Sambaran Bandyopadhyay}, \bibinfo{person}{Arnab Biswas}, \bibinfo{person}{Anlu Xing}, {and} \bibinfo{person}{Karthik Subbian}.} \bibinfo{year}{2022}\natexlab{}.
\newblock \showarticletitle{Shopping queries dataset: A large-scale ESCI benchmark for improving product search}.
\newblock \bibinfo{journal}{\emph{arXiv preprint arXiv:2206.06588}} (\bibinfo{year}{2022}).
\newblock


\bibitem[Rif’at et~al\mbox{.}(2018)]%
        {ner_Indonesian}
\bibfield{author}{\bibinfo{person}{Muhammad Rif’at}, \bibinfo{person}{Rahmad Mahendra}, \bibinfo{person}{Indra Budi}, {and} \bibinfo{person}{Haryo~Akbarianto Wibowo}.} \bibinfo{year}{2018}\natexlab{}.
\newblock \showarticletitle{Towards product attributes extraction in Indonesian e-commerce platform}.
\newblock \bibinfo{journal}{\emph{Computaci{\'o}n y Sistemas}} \bibinfo{volume}{22}, \bibinfo{number}{4} (\bibinfo{year}{2018}), \bibinfo{pages}{1367--1375}.
\newblock


\bibitem[Talmor et~al\mbox{.}(2018)]%
        {commonsenseqa}
\bibfield{author}{\bibinfo{person}{Alon Talmor}, \bibinfo{person}{Jonathan Herzig}, \bibinfo{person}{Nicholas Lourie}, {and} \bibinfo{person}{Jonathan Berant}.} \bibinfo{year}{2018}\natexlab{}.
\newblock \showarticletitle{Commonsenseqa: A question answering challenge targeting commonsense knowledge}.
\newblock \bibinfo{journal}{\emph{arXiv preprint arXiv:1811.00937}} (\bibinfo{year}{2018}).
\newblock


\bibitem[Wang et~al\mbox{.}(2024)]%
        {PairwiseFashion}
\bibfield{author}{\bibinfo{person}{Yu Wang}, \bibinfo{person}{Zexue He}, \bibinfo{person}{Zhankui He}, \bibinfo{person}{Hao Xu}, {and} \bibinfo{person}{Julian McAuley}.} \bibinfo{year}{2024}\natexlab{}.
\newblock \showarticletitle{Deciphering Compatibility Relationships with Textual Descriptions via Extraction and Explanation}. In \bibinfo{booktitle}{\emph{Proceedings of the AAAI Conference on Artificial Intelligence}}, Vol.~\bibinfo{volume}{38}. \bibinfo{pages}{9133--9141}.
\newblock


\bibitem[Yang et~al\mbox{.}(2024)]%
        {yang2024qwen2}
\bibfield{author}{\bibinfo{person}{An Yang}, \bibinfo{person}{Baosong Yang}, \bibinfo{person}{Binyuan Hui}, \bibinfo{person}{Bo Zheng}, \bibinfo{person}{Bowen Yu}, \bibinfo{person}{Chang Zhou}, \bibinfo{person}{Chengpeng Li}, \bibinfo{person}{Chengyuan Li}, \bibinfo{person}{Dayiheng Liu}, \bibinfo{person}{Fei Huang}, {et~al\mbox{.}}} \bibinfo{year}{2024}\natexlab{}.
\newblock \showarticletitle{Qwen2 Technical Report}.
\newblock \bibinfo{journal}{\emph{arXiv preprint arXiv:2407.10671}} (\bibinfo{year}{2024}).
\newblock


\bibitem[Young et~al\mbox{.}(2024)]%
        {young2024yi}
\bibfield{author}{\bibinfo{person}{Alex Young}, \bibinfo{person}{Bei Chen}, \bibinfo{person}{Chao Li}, \bibinfo{person}{Chengen Huang}, \bibinfo{person}{Ge Zhang}, \bibinfo{person}{Guanwei Zhang}, \bibinfo{person}{Heng Li}, \bibinfo{person}{Jiangcheng Zhu}, \bibinfo{person}{Jianqun Chen}, \bibinfo{person}{Jing Chang}, {et~al\mbox{.}}} \bibinfo{year}{2024}\natexlab{}.
\newblock \showarticletitle{Yi: Open foundation models by 01. ai}.
\newblock \bibinfo{journal}{\emph{arXiv preprint arXiv:2403.04652}} (\bibinfo{year}{2024}).
\newblock


\end{thebibliography}

\end{document}